\pdfoutput=1

\documentclass[11pt]{article}

\usepackage{emnlp2021}

\usepackage{times}
\usepackage{latexsym}

\usepackage{breqn}
\usepackage{amsmath}
\usepackage{url}

\usepackage[T1]{fontenc}

\usepackage[utf8]{inputenc}

\usepackage{microtype}
\usepackage{comment}

\usepackage{soul}
\usepackage{epstopdf}
\usepackage{booktabs}
\usepackage{graphicx}

\title{Examining Large Pre-Trained Language Models for Machine Translation: What You Don't Know About It}

\author{
          Lifeng Han$ ^1$, Gleb Erofeev$^{ 2}$,    Irina Sorokina$^{ 2}$,  Serge Gladkoff$ ^2$, 
\and \textbf{Goran Nenadic}$^{1}$ \\
         $^1$ The University of Manchester, UK \\ 
         $^2$ Logrus Global,  Translation \& Localization 
         \\ {\tt lifeng.han, g.nenadic@manchester.ac.uk} 
         \\
         {\tt 
         gleberof, irina.sorokina, serge.gladkoff@logrusglobal.com}         }
\begin{document}
\maketitle{}

\begin{abstract}
Pre-trained language models (PLMs) often take advantage of the monolingual and multilingual dataset that is freely available online to acquire general or mixed domain knowledge before deployment into specific tasks. 
Extra-large PLMs (xLPLMs) are proposed very recently to claim supreme performances over smaller-sized PLMs such as in machine translation (MT) tasks. 
These xLPLMs include Meta-AI's wmt21-dense-24-wide-en-X (2021) and NLLB (2022).
\textit{In this work, we examine if xLPLMs are absolutely superior to smaller-sized PLMs in fine-tuning toward domain-specific MTs. 
}
We use two different in-domain data of different sizes: commercial automotive in-house data and \textbf{clinical} shared task data from the ClinSpEn2022 challenge at WMT2022.
We choose popular Marian Helsinki as smaller sized PLM and two massive-sized Mega-Transformers from Meta-AI as xLPLMs.

Our experimental investigation shows that 
1) on smaller sized in-domain commercial automotive data, xLPLM wmt21-dense-24-wide-en-X indeed shows much better evaluation scores using S\textsc{acre}BLEU and hLEPOR metrics than smaller-sized Marian, even though its score increase rate is lower than Marian after fine-tuning; 
2) on relatively larger-size well prepared clinical data fine-tuning, the xLPLM NLLB \textbf{tends to lose} its advantage over smaller-sized Marian on two sub-tasks (clinical terms and ontology concepts) using \textit{ClinSpEn offered metrics} METEOR, COMET, and ROUGE-L, and totally lost to Marian on Task-1 (clinical cases) on \textit{all official metrics} including S\textsc{acre}BLEU and BLEU; 3) \textbf{metrics do not always agree} with each other on the same tasks using the same model outputs; 4) clinic-Marian ranked No.2 on Task-1 (via S\textsc{acre}BLEU/BLEU) and Task-3 (via METEOR and ROUGE) among all submissions.

\end{abstract}

\section{Introduction}

Owing to the recent development of neural machine translations (NMTs) \cite{kalchbrenner13emnlp,cho2014encoder-decoder,DBLP:journals/corr/BahdanauCB14,akhbardeh-EtAl:2021:WMT,han2022investigation}, 
especially the self-attention based Transformer learning structures \cite{devlin-etal-2019-bert,google2017attention}, pre-trained language models (PLMs) have been dominant in  natural language understanding (NLU) and natural language processing (NLP) tasks.
These applications include Long-Short Term Memory (LSTM) and BERT (Pre-training of Deep Bidirectional Transformers) based models to text mining \cite{2017neuroner,Han_Wu_etal2022_PLM4clinical}, question-answering \cite{dong-etal-2021-efficientbert-progressively_search}, reading comprehension \cite{schlegel2021emerging}, and summarisation \cite{perez-beltrachini-lapata-2021-models_summarisation}, etc., in addition to MT \cite{han-etal-2021-chinese,han2022overview_mte,han_gladkoff_metaeval_tutorial2022}.

PLMs often have a large amount of trainable parameters for downstream applications. For instance, in translation task, the popular
Marian NMT \cite{mariannmt} pre-trained by Microsoft Translator team \footnote{\url{https://translator.microsoft.com}} on OPUS \footnote{\url{https://opus.nlpl.eu}} \cite{tiedemann2012parallel_OPUS} multilingual corpus has 7.6 million parameters, which can still be fine-tuned on Google's Colab or AWS at virtually no cost. 
However, very recent work has shown much larger PLMs that have much more parameters than smaller models, e.g. the multi-lingual Transformer model  submitted to WMT2021 shared task by Meta-AI research group ``wmt21-dense-24-wide-en-x''(WMT21fb) \cite{tran2021facebook}, which has 4.7 billion parameters, i.e. 618 times bigger than Marian, and does not fit into regular GPUs. 
In this year, Meta-AI published another model NLLB \cite{https://doi.org/10.48550/arxiv.2207.04672_NLLB} that has 54.5 billion parameters and covers 200 languages in the full model.
From now on, we name both ``wmt21-dense-24-wide-en-x'' and NLLB as Meta-AI's \textit{Mega-Transformer} models.
Meta-AI's Mega-Transformer (WMT21fb) has claimed the best performing system on 10 out of 14 language pairs in WMT2021 shared task including winning bilingual-trained models.

In this work, we raise the question whether extra-large PLMs (xLPLMs) such as Meta-AI's Mega-Transformers have absolute superiority in NMT tasks on domain-specific fine-tuning.
We prepare experimental investigation on two different data set to answer this question.
One is our specific automotive domain in-house commercial data and the other is clinical domain data from ClinSpEn2022 challenge task we attended which is affiliated with WMT2022 \footnote{The 7th Conference on MT \url{https://www.statmt.org/wmt22/}}. 

We set up the following \textit{hypothesis} and \textit{research questions}.
Our \textit{\textbf{hypothesis}} is:
xLPLMs do not absolutely demonstrate superiority over smaller sized PLMs 
in NMT fine-tuning and it shall depends on specific tasks deployed including domain topic, size of available in-domain data, and performance-cost trad-off.

From this hypothesis we derive two \textit{research questions (\textbf{RQs})}: 
1) Do xLPLMs always demonstrate better performances in NMT over smaller sized PLMs for domain fine-tuning? 2) if not, in what situations? 

To the best of our knowledge, this is the first published work that has been carried out in the field on fine-tuning Meta-AI's extra-large multilingual PLM Maga-Transformers, and in translating specialised automotive and clinical data.

The rest of the paper is organised as below: 
Section \ref{section_related_work} introduces more details on related work to ours including PLMs and fine-tuning in automotive and clinical domains, 
Section \ref{section_model_setting} describes our initial model settings including deployed baseline models, 
Section \ref{section_experimental} presents our experimental evaluation carried out on our in-house commercial automotive domain data, 
Section \ref{section_wmt22} describes our system submission to ClinSpEn Biomedical-MT challenge task at WMT2022 on clinical data,
and Section \ref{section_discussion} gives our conclusion and future work plan.

\section{Related Work}
\label{section_related_work}
Fine-tuning PLMs has been in practice towards different domain applications in recent years. 
For instance, \newcite{DBLP:journals/corr/abs-2110-07477_2021LPLM4recom} carried out experimental investigation on fine-tuning  PLMs for conversational recommendation system, 
\newcite{chakraborty-etal-2020-biomedbert,10.1145/3458754_PubMedBERT,10.1093/bioinformatics/btz682_BioBERT,alsentzer-etal-2019-publicly_bioclinicalBERT} built biomedical and clinical domain pre-trained models using BERT structure and PubMed data on scientific publications,
and then
  \newcite{Han_Wu_etal2022_PLM4clinical,Han_etal2022_HealTAC22_TransformerCRF} developed new  machine learning structures using PLM Transformer and BERT as encoders in concatenation with statistical graph-based conditional random fields (CRFs) as decoders for clinical text mining.

However, aforementioned work did not deploy extra-large PLMs in a scale as Meta-AI's multilingual Mega-Transformers. 
For example, the PLMs (Transformer-CRFs) deployed by \newcite{Han_Wu_etal2022_PLM4clinical} as baseline have around 42 million of trainable parameters, which set is already relatively large, even though it is still far from Mega-Transformers' 4.7 billion and 54.5 billion parameters.

Regarding PLM applications in automotive domain,
the only recent work we found is from \newcite{romellmultilingual_2022msc_PLM4automotive}, who tested the Distil-BERT and XLM-RoBERTa PLMs for text classification task using Swedish truck manufacturer data, instead of MT.

There are also researchers working on the overview of model comparability, bench-marking, and fine-tuning methodologies regarding larger scale PLMs, e.g., from \newcite{assenmacher2021comparability_phd_PLM,ruder2021lmfine-tuning}.

Overall, none of the work mentioned before has investigated into extra-large Mega-Transformer level PLMs (xLPLMs) for NMT in automotive and clinical domains, especially their comparisons to smaller sized PLMs.

\section{Initial Model Settings}
\label{section_model_setting}
To investigate into PLMs with fine-tuning for specialised domain NMT from different scales, we firstly deploy two of such models in different sizes from a multilingual setting. 
The first one is the popular Marian NMT model developed in C++ since 2018 using deep RNN and Transformer \cite{mariannmt}.   
It is mostly maintained by the Microsoft Translator team and 
features with efficiency, fast training, and state-of-the-art NMT architectures \footnote{available at \url{https://marian-nmt.github.io}}. This PLM has a smaller sized 7.6 million trainable parameters.

The second one we use for fine-tuning is one of the extra-large PLMs (\textit{xLPLMs}) Meta-AI's Mega-Transformer ``wmt21.dense-24-wide.En-X'' \cite{tran2021facebook} developed for WMT2021 shared task on multilingual MT, which was submitted to 14 language pairs and claimed the best on 10 of them \footnote{package ``wmt21.dense-24-wide.En-X'' available at \url{https://github.com/facebookresearch/fairseq/tree/main/examples/wmt21}}. It has 4.7 billion trainable parameters, which is super large in comparison to Marian model.
In the later section (\ref{section_wmt22}), we will explain another Mega-Transformer Model NLLB developed in this year by Meta-AI and deploy it for our ClinSpEn2022 shared task submission on clinical domain.

\section{Model Fine-Tuning and Comparison on Commercial Automotive Data}
\label{section_experimental}

\subsection{In-house Corpus and Hardware}

At the development stage, we use our in-house prepared domain-specific commercial corpus from automotive field. 
We split our data set into 90\% vs 10\% for fine-tuning and testing respectively and make sure that the test data is not seen during the fine-tuning / development stage \footnote{Because this is a commercial corpus, we do not give much details on it but this does not affect the experimental findings we achieved}. 
We use a larger GPU from NVIDIA A100 with 80GB VRAM for our experiments because of the much higher computational powers the Mega-Transformer model requires.

\subsection{Our Evaluation Setup}

BLEU \cite{papineni-etal-2002-bleu} has always been criticised by researchers on its reliability. This includes very recent work by \newcite{google2021human_evaluation_TQA}, which demonstrates that BLEU has closer correlation to lower quality crowd sourced human evaluation then to expert based human evaluation, and by \newcite{han-etal-2021-chinese}, which investigation on Chinese-English NMT shows that BLEU score fails to reflect the real quality differences between NMT systems especially on translating multi-word expressions (MWEs) and terms \cite{han-etal-2020-alphamwe}.

Furthermore, BLEU scores can be very different caused by configurations, such as tokenisation and normalisation strategies applied to the reference text which can lead to 1.8 margin of difference reported by \cite{post-2018-call4clarity}. 
In light of these findings, we adopt two alternative evaluation metrics, i.e. S\textsc{acre}BLEU \cite{post-2018-call4clarity} and hLEPOR \cite{han2013hLEPOR_MTsummit,cushLEPOR21MTsummit,han2021cushlepor} that we will give further details about.

\subsubsection{Revisiting S\textsc{acre}BLEU}

S\textsc{acre}BLEU is developed by the work from \newcite{post-2018-call4clarity} and is maintained online in its Python version \footnote{available at \url{https://github.com/mjpost/sacrebleu}}. The author discussed the uncertainty regarding reporting BLEU scores by MT researchers. 
This is involved in many parameter settings when using BLEU metric including number of references, length penalty computation on multi-references, maximum n-gram, and smoothing applied to 0-count n-grams. 
Because of such variations, when MT researchers report the BLEU scores from their system, ``the BLEU'' score actually cannot be reproduced in many cases due to lack of detailed technical description of encoder, etc.
.

To address these issues, S\textsc{acre}BLEU added some constrains while using BLEU metric. These include the applying of its own metric-internal pre-processing for detokenised system outputs, the avoiding of user handling reference set via automatically downloading from WMT, and the export of a summary on settings used.

\subsubsection{Revisiting hLEPOR}
\label{revisit_hLEPOR}

hLEPOR is an augmented metric for automatic MT evaluation which was firstly proposed in WMT2013 Metrics shared task \cite{han-etal-2013wmt-description,han2013hLEPOR_MTsummit} and was reported as one of the best performing metrics at both system level \cite{machacek-bojar-wmt2013-results} and segment level \cite{DBLP:conf/naacl/GrahamBM15} \footnote{The python version is available at \url{https://pypi.org/project/hLepor/} and the original Perl code at \url{https://github.com/poethan/LEPOR}}. 
It is calculated via a weighted harmonic mean of several main factors including sentence length penalty, position difference penalty, precision, and recall. Furthermore, there are more weighting parameters among all the sub-factors. 
Let's see the brief formulas below:

\begin{gather*} 
\text{\textit{h}LEPOR} = {Harmonic(w_{LP}LP},
\\ w_{NPosPenal}NPosPenal, w_{HPR}HPR)
\end{gather*}

\noindent
where \textit{LP} is the sentence length penalty factor and is calculated as:

\begin{align*}
    \text{LP}&= \left\{
                \begin{array}{ll}
                  e^{1-\frac{Length_{ref}}{Length_hyp}} \; if\; Length_{hyp} < Length_{ref}\\
                  1 \; if \; Length_{hyp} = Length_{ref} \\
                  e^{1-\frac{Length_hyp}{Length_{ref}}} \; if \; Length_{hyp} > Length_{ref}
                \end{array}
              \right.
\end{align*}

Then, n-gram based position difference penalty (NPD) is used to measure the word position and order difference among matched words between system output and reference translation ($MatchN_{hyp}$ and $MatchN_{ref}$). 

\begin{gather*} 
NPosPenal = e^{-NPD}
\\ NPD = \frac{1}{Length_{hyp}}\sum_{i=1}^{Length_{hyp}}|PD_i|
\\ |PD_i|= |MatchN_{hyp}-MatchN_{ref}|
\end{gather*}

\noindent Finally, the weighted harmonic mean of precision and recall is calculated using this formula.

\begin{gather*} 
HPR = \frac{(\alpha+\beta)Precision x Recall}{\alpha Precision + \beta Recall}
\\ Precision = \frac{Aligned_{num}}{Length_{hypothesis}}
\\ Recall = \frac{Aligned_{num}}{Length_{reference}}
\end{gather*}

\noindent hLEPOR is an extended version of the original LEPOR metric \cite{han2012lepor,han2014lepor}. hLEPOR also has a latest customised version named cushLEPOR which uses automatic hyper-parameter optimisation framework Optuna \cite{Optuna2019kdd} to achieve better and easier feature weights fine-tuning towards specific language pairs and domains in practice. 
It was reported as one of the best performing metrics in WMT2021 \cite{cushLEPOR21MTsummit,han2021cushlepor}
on the officially-ranked language pairs English-German and Chinese-English on News domain, and English-Russian on TED talk data \cite{freitag-etal-2021metrics-findings} where human expert level evaluations were available.
hLEPOR is also gaining popularity in other NLP task evaluations, e.g. language generation (NLG) \cite{NLG_evaluation2017emnlp_Novikova_etal,arxiv2021NLG,Marzouk2021CNL4mt}, language understanding (NLU) \cite{2021NLU_ruder_multilingual_evaluation}, text summarization (ATS) \cite{bhandari-etal-2020-evaluating_summarization}, and searching \cite{search_eval2021Liu}.

\subsection{Evaluation Results}

The evaluation scores using S\textsc{acre}BLEU and hLEPOR are shown in Table \ref{tab:SacreBLEU_score_comp} and \ref{tab:hLEPOR_score_comp} respectively.
From Table \ref{tab:SacreBLEU_score_comp}, we can see that the fine-tuning has successfully improved each single n-gram precision score in S\textsc{acre}BLEU for both Marian and Mega-Transformer models, leading to an overall 150.14\% and 75.81\% score increasing.
Similarly, Table \ref{tab:hLEPOR_score_comp} shows that our in-domain fine-tuning improved hLEPOR scores on Marian and Mega-Transformer models via 32.16\% and 26.01\%. 

Like BLEU, S\textsc{acre}BLEU is precision based metric. The very large margin evaluation score increases in S\textsc{acre}BLEU (150.14\% and 75.81\%) indicates that according to reference translation, our fine-tuned models produce more fluent output than the baseline in this domain specific test set.
Unlike S\textsc{acre}BLEU, hLEPOR is an augmented metric with comprehensive factors, including recall and positional difference penalty, in addition to precision. 
The large margins of hLEPOR score increase, i.e. 32.16\% and 26.01\% tell that the fine-tuned models can also have more adequate translation towards this domain, in addition to maintaining higher fluency.

In summary, the fine-tuning of these two PLMs has demonstrated evaluation score improvement with large margins in commercial domain data. xLPLM Mega-Transformer has much higher S\textsc{acre}BLEU evaluation score than Marian before fine-tuning, 39.12 vs 19.64, which indicates its larger amount of knowledge acquired. 
However, after fine-tuning, the S\textsc{acre}BLEU scores of them are much closer, 50.33 vs 45.20. 
This means that fine-tuning of smaller sized PLM for this commercial data is far more effective than the xLPLM Mega-Transformer from computation and time cost point of view, as well as the cost of computational power itself, since supercomputer time is much more expensive.

This partially verifies our assumption that xLPLMs do not always win smaller sized PLMs in practical applications when computational cost is in place and when time is constrained.

To further investigate our research questions, we carry out another experimental evaluation on clinical domain data via attending the ClinSpEn2022 shared task challenge which will be detailed in the next section.

\begin{table*}[!t]
\begin{center}
\centering
\begin{tabular}{crccccc}
\toprule
\multicolumn{1}{c}{} 
     & \multicolumn{6}{c}{Marian }    \\ \hline 
\multicolumn{1}{c}{} 
     & \multicolumn{1}{c}{uni-gram }     
                & bi-gram & tri-gram & 4-gram & BP & Overall\\
\midrule
Before fine-tuning & 19.64  & 10.96 & 4.56 & 2.00 & 1.0 & 7.38 \\
After fine-tuning &  45.20 & 24.54 & 14.44 & 8.69 & 0.96 & 18.46 ($\uparrow$150.14\%)  \\
\hline
\multicolumn{1}{c}{} 
     & \multicolumn{6}{c}{Mega-Transformer (wmt21fb) }    \\ \hline 
\multicolumn{1}{c}{} 
     & \multicolumn{1}{c}{uni-gram }     
                & bi-gram & tri-gram & 4-gram & BP & Overall\\
\midrule
Before fine-tuning & 39.12  & 18.81 & 9.78 & 5.23 & 1.0 & 13.93 \\
After fine-tuning & 50.33  & 30.14 & 19.47 & 12.85 & 0.99 & 24.49 ($\uparrow$75.81\%) \\
\bottomrule
\end{tabular}
\caption{S\textsc{acre}BLEU score comparisons on the MT test set: before vs after fine-tuning}
\label{tab:SacreBLEU_score_comp}
\end{center}
\end{table*}

\begin{table}[!t]
\begin{center}
\centering
\begin{tabular}{crc}
\toprule
\multicolumn{1}{c}{} 
     & \multicolumn{1}{c}{Marian }     
                & Mega-Transformer\\
\midrule
Before fine-tuning & 36.91    & 47.55  \\
After fine-tuning & 48.78    & 59.92  \\
Rate($\uparrow$) & 32.16\%    & 26.01\%  \\
\bottomrule
\end{tabular}
\caption{hLEPOR score comparisons on the MT test set: before vs after fine-tuning}
\label{tab:hLEPOR_score_comp}
\end{center}
\end{table}

\section{Submission to ClinSpEn at WMT22}
\label{section_wmt22}
In this section, we introduce our system submissions to Biomedical-MT task in WMT2022. In this task, we attended the affiliated clinical domain machine translation on Spanish-English language pair (ClinSpEn) task \footnote{\url{https://temu.bsc.es/clinspen/}}, which is hosted in CodaLab \cite{codalab_competitions} \footnote{\url{https://codalab.lisn.upsaclay.fr/competitions/6696}}. 

The aim of this task is to promote the development of MT models on medical domain via three sub-tasks: 1) Clinical Cases (CC): on 202 COVID-19 clinical case reports; 2) Clinical Terms (CT): using more than 19K parallel terms extracted from biomedical literature and electric health records (EHRs); 3) Ontology Concepts (OC): using more than 2K parallel concepts from biomedical ontology. 
The translation direction on these three sub-tasks are EN$\rightarrow$ES, EN$\leftarrow$ES, and EN$\rightarrow$ES respectively.

\subsection{Corpus Used}
In addition to the official corpora prepared by the ClinSpEn organisers, we used some external corpora for our model fine-tuning. 
This is because that neural-network based machine learning models are data dependent while the officially offered parallel sample sentences are very limited.
We found useful biomedical Spanish-English corpora described in \cite{neveol-etal-2018-parallel_biomed} from WMT\footnote{\url{https://github.com/biomedical-translation-corpora}}, and MeSpEn corpora from \cite{villegas2018mespen}\footnote{\url{https://zenodo.org/record/3562536}}, which include 
Spanish Bibliographical Index in Health Sciences (IBECS), 
Scientific Electronic Library Online (SciELO), and U.S. National Library of Medicine (PubMed and MedlinePlus).
However, due to the time restriction for this shared task, we only managed to get 250,000 aligned pairs from IBECS after careful preparation, which is a bibliographical data collecting scientific articles from different fields of health sciences, maintained by the Spanish National Health Sciences Library.

\subsection{Adaptations on xLPLM: NLLB}
Two systems we submitted to ClinSpEn2022 are clinic-Marian and clinic-NLLB \cite{https://doi.org/10.48550/arxiv.2207.04672_NLLB}. We reported our clinic-WMT21fb model outputs in a followup work \cite{Han_etal_MMPLM_ZeroShotNMT2022} (also due to the time restriction). 
Some training parameters and training logs for clinic-Marian are listed below:

\begin{itemize}
    \item batch size = 64
    \item gradient accumulation steps = 1
    \item weight decay = 0.01
    \item learning rate = 2e-5
    \item number of training epochs = 1
    \item number of examples = 225,000
\end{itemize}

NLLB (No Language Left Behind) is another extra-large PLM model built by Meta-AI freshly in this year \footnote{The project page \url{https://ai.facebook.com/research/no-language-left-behind/}}, which was targeting low-resource languages via knowledge transfer from high-resource ones, and Spanish is among the high-resource languages covered by NLLB \footnote{Models available at \url{https://huggingface.co/docs/transformers/model_doc/nllb}}. 
NLLB-200 has a total of 54.5 billion parameters in its full model as the authors mentioned. 
In this shared task, we applied the distilled version of NLLB, i.e. the ``NLLB-200-distilled-1.3B'' which still has 1.3 billion trainable parameters \footnote{\url{https://huggingface.co/facebook/nllb-200-distilled-1.3B}}. 
As Meta-AI's ``wmt21.dense-24-wide.en-X'' model we used in the earlier section, we call NLLB-distilled as one of their \textit{Mega-Transformers}.

Some fine-tuning parameters for NLLB-distilled are listed below:
\begin{itemize}
    \item batch size = 24
    \item gradient accumulation steps = 8
    \item weight decay = 0.01
    \item learning rate = 2e-5
    \item number of training epochs = 1
    \item encoder-decoder layers = 24+24
\end{itemize}

The fine-tuned clinic-NLLB model has relatively apparent evaluation score increase using S\textsc{acre}BLEU in comparison to baseline model on both translation directions, as shown in Table \ref{tab:SacreBLEU_score_NLLB_dev}, for EN$\rightarrow$ES and ES$\rightarrow$EN in the upper and middle parts of the table with increasing rate 11.74\% and 9.70\% respectively. This demonstrates that that fine-tuning was successful.

Interestingly, if we fine-tune the model in one direction and carry out the inference translation in the opposite direction, the model performance will have a big drop even though it is the same language pair. 
This tells that pre-trained LMs lose their generalisation after fine-tuning. For instance, in the bottom of Table \ref{tab:SacreBLEU_score_NLLB_dev}, we demonstrate that if the model is fine-tuned in English-to-Spanish direction and  the inference test is carried out in Spanish-to-English direction, the overall S\textsc{acre}BLEU score has a 14.37\% drop in comparison to without fine-tuning. 
So, we carried out fine-tuning on both translation directions for the system submission to three sub-tasks at ClinSpEn2022.

\begin{table*}[!t]
\begin{center}
\centering
\begin{tabular}{crccccc}
\toprule
\multicolumn{1}{c}{} 
     & \multicolumn{6}{c}{English-to-Spanish (tune+test) }    \\ \hline 
\multicolumn{1}{c}{} 
     & \multicolumn{1}{c}{uni-gram }     
                & bi-gram & tri-gram & 4-gram & BP & Overall\\
\midrule
Before fine-tuning & 65.93  & 45.51 & 33.71 & 25.44 & 1.0 & 40.05 \\
After fine-tuning &  70.25 & 50.58 & 38.78 & 30.17 & 0.99 & 44.75 ($\uparrow$11.74\%)  \\
\hline
\multicolumn{1}{c}{} 
     & \multicolumn{6}{c}{Spanish-to-English (tune+test) }    \\ \hline 
\multicolumn{1}{c}{} 
     & \multicolumn{1}{c}{uni-gram }     
                & bi-gram & tri-gram & 4-gram & BP & Overall\\
\midrule
Before fine-tuning & 65.36& 42.54& 30.58& 22.60& 1 &37.23 \\
After fine-tuning & 68.51& 46.27& 34.07& 25.76 & 1 & 40.84 ($\uparrow$9.70\%) \\ \hline\hline 
\multicolumn{1}{c}{} 
     & \multicolumn{6}{c}{English-to-Spanish (tune) \& Spanish-to-English (test) }    \\ \hline 
\multicolumn{1}{c}{} 
     & \multicolumn{1}{c}{uni-gram }     
                & bi-gram & tri-gram & 4-gram & BP & Overall\\
\midrule
Before fine-tuning (es2en) &  65.36& 42.54& 30.58& 22.60& 1 &37.23 \\
After \textit{Reverse} fine-tuning & 58.17& 36.48& 25.85& 18.84 &1.0 & 31.88 ($\downarrow$14.37\%)  \\
\hline
\bottomrule
\end{tabular}
\caption{S\textsc{acre}BLEU score comparisons using NLLB: baseline vs fine-tuned in clinical domain.}
\label{tab:SacreBLEU_score_NLLB_dev}
\end{center}
\end{table*}

\subsection{Official Evaluation Metrics}
The official evaluation metrics used by CinSpEn2022 shared task are METEOR \cite{BanerjeeLavie2005}, S\textsc{acre}BLEU \cite{post-2018-call4clarity}, COMET \cite{rei-etal-2020-comet}, BLEU-HF (HuggingFace) \cite{papineni-etal-2002-bleu}, and ROUGE-L-F1 \cite{lin-2004-rouge}. Among these, METEOR is a metric using both precision and recall not only on word surface level but also introducing paraphrasing features. COMET was proposed recently by taking advantage of cross-lingual PLMs using knowledge from both source and target languages. ROUGE was originally designed for text summarisation evaluation using n-gram co-occurrences, while ROUGE-L
added the Longest Common Sub-sequence (LCS) feature from translation study.

\subsection{Evaluation Scores on Three Tasks}

We present the MT evaluation scores using five official metrics through CodaLab platform on the three sub-tasks in Table \ref{tab:clinSpEn_eval_score_t123}, for translating clinical cases, clinical terms, and ontology concepts. 
The two fine-tuned models are clinic-Marian and clinic-NLLB (one of the Mega-Transformers). 
From this shared task evaluation outcomes, the xLPLM clinic-NLLB starts to lose its comparisons to far smaller-sized clinic-Marian in Task-2 (CT) and 3 (OC), especially on METEOR and ROUGE-L scores but also on COMET (OC). What is very noticing is that clinic-Marian has an overall win on Task-1 (CC) via all evaluation metrics.

From the evaluation results on Task 2 and 3, i.e. CT and OC, we can see that the evaluation metrics do not agree with each other always.
For instance, clinic-Marian wins METEOR and ROUGE-L on Task 2 but loses on other metrics, while clinic-NLLB wins S\textsc{acre}BLEU and BLEU-HF on Task 3 but loses on other metrics. 
This phenomenon is very interesting which tells that variation metrics from BLEU including BLEU-HF and S\textsc{acre}BLEU tend to not agree with other metric families including METEOR, COMET, and ROUGE-L. 
Furthermore, the same metric does not always agree with itself on different tasks, or the two MT models perform differently across tasks. For instance, COMET score says clinic-Marian and clinic-NLLB wins Task 3 (0.9495) and 2 (1.0290) respectively.
Due to the time restriction from this shared task and the limited computational resource we have, our second model (clinic-NLLB) was submitted after the official deadline.

This experimental investigation shows that with carefully prepared and larger amount of domain specific data for fine-tuning, the xLPLMs tend to lose its advantage over smaller sized PLMs using several automatic metrics. Thus it further verifies our  hypothesis and research questions.

\begin{table*}[!t]
\begin{center}
\centering
\begin{tabular}{crcccc}
\toprule
\multicolumn{1}{c}{} 
     & \multicolumn{5}{c}{clinic-Marian }    \\ \hline 
\multicolumn{1}{c}{MT} 
     & \multicolumn{1}{c}{S\textsc{acre}BLEU}     
                & METEOR & COMET & BLEU-HF & ROUGE-L-F1 \\
\midrule
Task-I: clinical cases & \textit{38.18} &\textit{0.6338} &\textit{0.4237} &\textit{0.3650} &\textit{0.6271}  \\
Task-II: clinical terms & 26.87 &\textit{0.5885} &0.9791 &0.2667 &\textit{0.6720} \\
Task-III:clinical concepts &39.10 & \textit{0.6262} &\textit{0.9495} & 0.3675 &\textit{0.7688} \\
\hline
\multicolumn{1}{c}{} 
     & \multicolumn{5}{c}{clinic-NLLB (Mega-Transformers) }    \\ \hline 
\multicolumn{1}{c}{MT} 
     & \multicolumn{1}{c}{S\textsc{acre}BLEU}     
                & METEOR & COMET & BLEU-HF & ROUGE-L-F1 \\
\midrule
Task-I: clinical cases & 37.74 & 0.6273& 0.4081& 0.3601& 0.6193 \\
Task-II: clinical terms  & \textit{28.57} & 0.5873 & \textit{1.0290} & \textit{0.2844} & 0.6710 \\
Task-III: ontology concepts & \textit{41.63}  & 0.6072& 0.9180& \textit{0.3932} & 0.7477 \\
\bottomrule
\end{tabular}
\caption{Evaluation Scores using Official CodaLab Platform from ClinSpEn2022 Benchmark on Fine-tuned Models. \textit{italic} scores indicate winner on the specific task using the specific metric (last digit rounded).}
\label{tab:clinSpEn_eval_score_t123}
\end{center}
\end{table*}

\begin{table*}[!t]
\begin{center}
\centering
\begin{tabular}{crcccc}
\toprule
\multicolumn{1}{c}{} 
     & \multicolumn{5}{c}{Task-1: Translating Clinical Cases}    \\ \hline 
\multicolumn{1}{c}{Teams} 
     & \multicolumn{1}{c}{S\textsc{acre}BLEU}     
                & METEOR & COMET & BLEU & ROUGE \\
\midrule
DtranX & \textit{41.06} & {0.6633} & \textit{0.4610} & \textit{0.3926} &\textit{{0.6490}} \\
Logrus-UoM (ours) & {\underline{38.17}} & {0.6337} & {0.4237} & \underline{0.3650} & {0.6270}  \\
Optum(run4) & 38.12 & {0.6447} &0.4425 &0.3642 & {0.6285} \\
Avellana Translation & 36.64 & \textit{0.6637} & {0.3920} & 0.3519 &{0.6333} \\
\hline
\multicolumn{1}{c}{} 
     & \multicolumn{5}{c}{Task-3: Translating Ontology Concepts}    \\ \hline 
\multicolumn{1}{c}{Teams} 
     & \multicolumn{1}{c}{S\textsc{acre}BLEU}     
                & METEOR & COMET & BLEU & ROUGE \\
\midrule
DtranX & \textit{58.24} & \textit{0.6275} & \textit{1.2496} & 0.5724 &\textit{{0.7839}} \\
Optum(run4) & 44.97 & 0.5880 & 1.1197 &0.4396 & 0.7479 \\
Logrus-UoM (ours) & {{39.10}} & \underline{0.6261} & {0.9494} & {0.3674} & \underline{0.7688}  \\
Avellana Translation & 31.72 & {0.5707} & 0.3841 & 0.3042 & 0.7621 \\
\hline
\bottomrule
\end{tabular}
\caption{Comparisons on Task 1 and 3 across teams (ranked via S\textsc{acre}BLEU chosen by the organisers).}
\label{tab:clinSpEn_eval_score_t1_offcial}
\end{center}
\end{table*}

\subsection{Comparisons to Other Teams}
In the officially valid submissions (before the shared task deadline ended) for three tasks, there are four teams for Task-1 and Task-3 including Avellana Translation, DtranX, Optum and ours \footnote{\url{https://statmt.org/wmt22/biomedical_results.pdf} }. In addition to these four teams, Task-2 has another team Huawei, making it in-total five teams. Optum and Huawei have both multiple submissions/runs while other teams submitted one run.
Our submission clinic-Marian ranked number 2 in both Task-1 and Task-3 via S\textsc{acre}BLEU/BLEU and METEOR/ROUGE respectively, as in Table \ref{tab:clinSpEn_eval_score_t1_offcial} \underline{underlined}.
There are four runs from Optum team for both Task-1/3 and single submission by other teams. Table \ref{tab:clinSpEn_eval_score_t1_offcial} includes the best submission from Optum. 
There is a little difference in the last digit of the evaluation scores between our own record (Table \ref{tab:clinSpEn_eval_score_t123}) and the official record (Table \ref{tab:clinSpEn_eval_score_t1_offcial}), which is because that we rounded the last digit scores while the official ones did not.
This result shows that metrics tend to not agree with each others in many cases. For instance, on Task-3, our clinic-Marian has very similar score to DtranX on METEOR (0.6261 vs 0.6275) only from the third digit which is a metric using paraphrase and semantic similarity features; however, the score difference on BLEU is so large (39.10 vs 58.24) via S\textsc{acre}BLEU which rises the issue again on the credibility of BLEU metric. 
There are not many teams submitting their results into this clinical domain machine translation task in comparison to the traditional news domain MT task,
which indicates that it is still a relatively new domain and calls for more attentions from MT researchers in the future.

\section{Discussion and Future Work}
\label{section_discussion}

In this work, we carried out experimental investigations on if extra-large pre-trained language models (PLMs) always demonstrate superiority over much smaller-sized PLMs using two domain specific data. 
The first experimental results using Marian vs ``wmt21.dense-24-wide.En-X'' shows that even though xLPLM still perform better evaluation scores in comparison to much smaller sized Marian, their score difference is much smaller after fine-tuning and the xLPLM costs more than smaller PLM from performance-cost trade-off point of view in practical applications, e.g. for language service providers (LSPs). 
The second experimental results using clinical data show that with carefully prepared certain amount of fine-tuning data (250k sentence pairs), the xLPLM NLLB even loses with its evaluation score in comparison to smaller PLM Marian in Task 1 ``clinical cases'' over all automatic metrics used, and in Task 2 ``clinical terms'' and 3 ``ontology concepts'' on partial of the automatic evaluation metrics officially used by ClinSpEn2022.
Finally, our system submission clinic-Marian ranked the second place using S\textsc{acre}BLEU/BLEU for Task-1, and using METEOR/ROUGE for Task-3 among all teams who submitted on-time before the shared task deadline.

We looked into the translation outputs from clinic-NLLB for error analysis, and it shows that some of the translation errors come from very literal translation, and others come from gender related mistakes. 
In conclusion, \textit{our two stage experimental investigations verify our hypothesis and RQs from different aspects}.

We also doubt if the official automatic metrics used for ClinSpEn challenge can correctly distinguish the NMT systems because mostly they do not really measure the translation output quality but the similarity to the gold standard single reference. Therefore, domain specific automatic evaluation metrics or metrics better measuring semantic similarities might be needed.  

In the future work, we plan to carry out more experimental investigations from qualitative aspects looking into translation errors using human experts and classifying them into possible categories with examples and statistics, especially from clinical domain.
This will allow us to validate automatic metrics with professional human judgements for this domain.

We will continue to fine-tune our models towards different domains and languages and use more of the available corpus for current clinical domain challenge task.
We also plan to try different state-of-the-art pre-trained language models for evaluation.

\section*{Acknowledgements}
The authors thank the ClinSpEn2022 shared task organisers for preparing the data set and evaluation platforms, thank Darryl Estrada for in communication with us during the competition.
The authors thank the project support from HIPS (R126035 task A05) and from JigSaw (R124782 task A07) at The University of Manchester. 
We thank the open research projects Marian Helsinki NMT and Meta-AI's wmt21.dense-24-wide.En-X (2021) and NLLB (2022) we used.

\bibliography{anthology,custom}
\bibliographystyle{acl_natbib}

\section*{Appendix}
More training logs from clinic-Marian:

\begin{itemize}
    \item global step = 3516
    \item training loss = 1.2236216656855212
    \item train runtime = 1945.9989
    \item train samples per second = 115.622
    \item trian steps per second = 1.807
    \item total flos = 2947034863632384.0
\end{itemize}

\noindent Parameters reported by S\textsc{acre}BLEU: 

\begin{itemize}
    \item lowercase = Ture
    \item tokenize = 13a
\end{itemize}

\end{document}